\documentclass[sigconf,screen]{acmart}
\usepackage{multirow}
\usepackage[ruled,vlined]{algorithm2e}
\AtBeginDocument{%
  }
\renewcommand\footnotetextcopyrightpermission[1]{}

\acmConference[MM'33]{Proceedings of the 33th ACM International Conference on Multimedia}{October 27--31, 2025}{Dublin, Ireland}
\acmBooktitle{Proceedings of the 33th ACM International Conference on Multimedia(MM'33), October 27--31, 2025, Dublin, Ireland}
\begin{document}

\title{Think Hierarchically, Act Dynamically: Hierarchical Multi-modal Fusion and Reasoning for Vision-and-Language Navigation}

\author{Junrong Yue}
\affiliation{%
  \institution{City University of Hong Kong, Dongguan Campus}\country{China}
}

\author{Chuan Qin}
\affiliation{%
  \institution{The University of Melbourne}\country{Australia}
}

\author{Bo Li}
\affiliation{%
  \institution{Tsinghua University \& Baidu Inc.}\country{China}
}

\author{Wenxin Zhang}
\affiliation{%
  \institution{University of Chinese Academy of Science}\country{China}
}

\author{Xinlei Yu}
\affiliation{%
  \institution{National University of Singapore}\country{Singapore}
}

\author{Xiaomin Lie}
\affiliation{%
  \institution{City University of Hong Kong, Dongguan Campus}\country{China}
  }
%   \city{Dongguan}
%   \state{Guangdong}
% \email{72405446@cityu-dg.edu.cn}

\author{Zhendong Zhao}
\affiliation{%
  \institution{University of Chinese Academy of Science}\country{China}
}

\author{Yifan Zhang}
\authornote{Corresponding author.}
%%\email{72405446@cityu-dg.edu.cn}
%%\orcid{1234-5678-9012}
%%\author{G.K.M. Tobin}
%%\email{webmaster@marysville-ohio.com}
\affiliation{%
  \institution{City University of Hong Kong, Dongguan Campus}\country{China}
  }

\begin{abstract}
  Vision-and-Language Navigation (VLN) aims to enable embodied agents to follow natural language instructions and reach target locations in real-world environments. While prior methods often rely on either global scene representations or object-level features, these approaches are insufficient for capturing the complex interactions across modalities required for accurate navigation. In this paper, we propose a Multi-level Fusion and Reasoning Architecture (MFRA) to enhance the agent's ability to reason over visual observations, language instructions and navigation history. Specifically, MFRA introduces a hierarchical fusion mechanism that aggregates multi-level features—ranging from low-level visual cues to high-level semantic concepts—across multiple modalities. We further design a reasoning module that leverages fused representations to infer navigation actions through instruction-guided attention and dynamic context integration. By selectively capturing and combining relevant visual, linguistic, and temporal signals, MFRA improves decision-making accuracy in complex navigation scenarios. Extensive experiments on benchmark VLN datasets including REVERIE, R2R, and SOON demonstrate that MFRA achieves superior performance compared to state-of-the-art methods, validating the effectiveness of multi-level modal fusion for embodied navigation.
\end{abstract}



\maketitle



\section{Introduction}
% 第一段没啥大问题，讲一下VLN任务是什么，然后为什么重要
Vision-and-Language Navigation (VLN) \cite{anderson2018visionandlanguagenavigationinterpretingvisuallygrounded,guhur2021airbertindomainpretrainingvisionandlanguage,qi2020objectandactionawaremodelvisual,qi2020reverieremoteembodiedvisual,zhu2021soonscenarioorientedobject,zhu2022diagnosingvisionandlanguagenavigationreally} represents a key research area in embodied Artificial Intelligence (AI), where autonomous agents are tasked with processing natural language instructions, interpreting dynamic 3D surroundings and performing accurate navigation actions. This capability is essential for applications ranging from assistive robotics to augmented reality, as it bridges the gap between high-level linguistic commands and low-level spatial reasoning. Unlike conventional navigation systems \cite{Schmidt_2019,Chow_2018,kalita2018pathplanningnavigationinside,Chow_2017} that rely solely on visual inputs or predefined waypoints, VLN demands seamless integration of multi-modal inputs, including visual perception, linguistic comprehension, and temporal context, to achieve robust performance in real-world scenarios. The complexity of VLN lies in its requirement for agents to ground abstract language concepts into concrete visual and spatial representations, making it a challenging but critical area of study.

\begin{figure}[h]
  \centering
  \includegraphics[width=\linewidth]{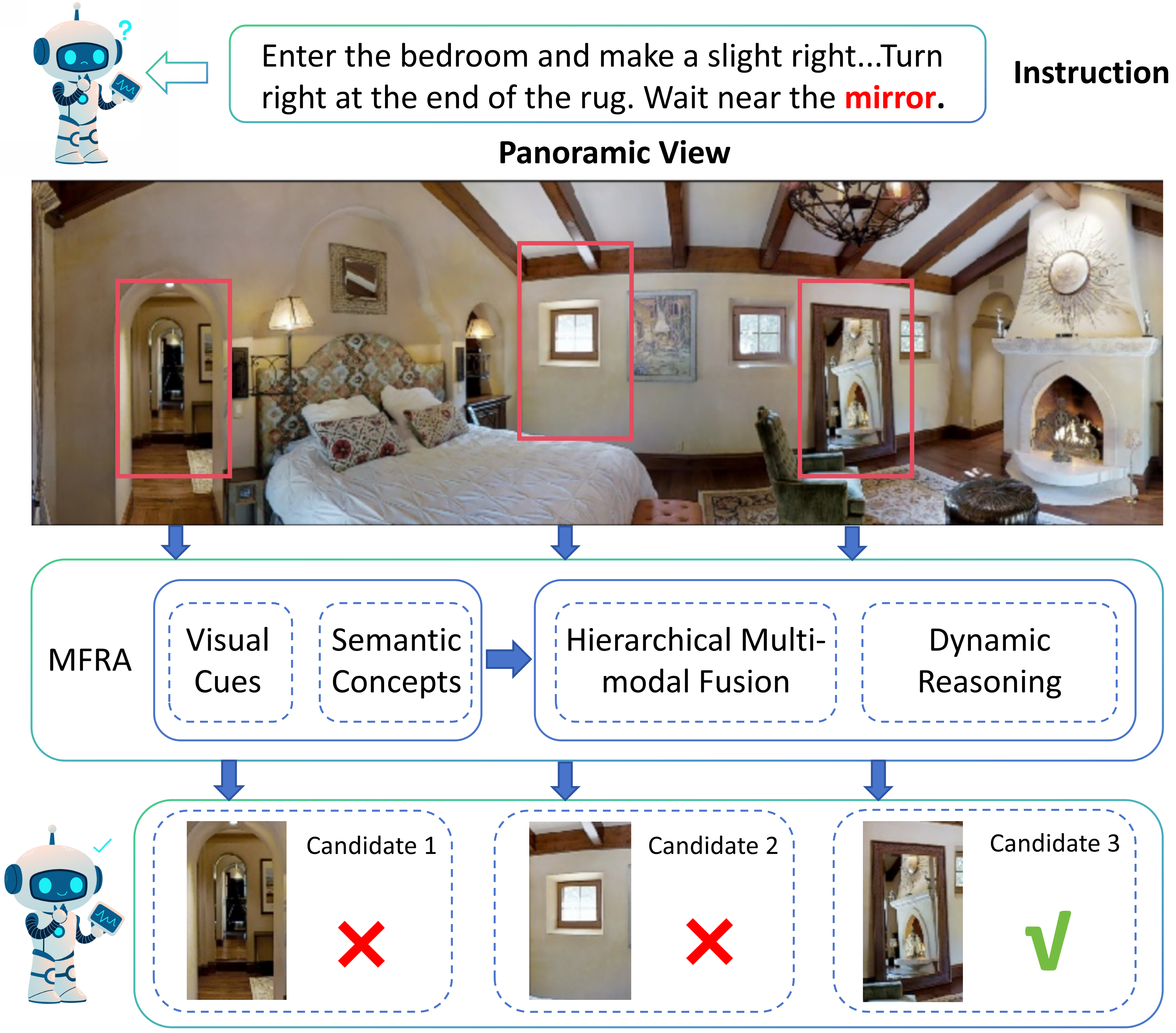}
  \caption{Illustration of MFRA selected navigable candidates,which provides crucial information such as attributes and relationships between objects for further VLN reasoning module. Best viewed in color.}
  % \vspace{-0.2cm}
  \label{fig:motivation}
\end{figure}

% 第二段讲传统做法是怎么做的，有什么问题，为什么会有这些问题 （二选一）
Early VLN approaches adopted polarized representation strategies, either relying exclusively on global scene embeddings \cite{chen2023historyawaremultimodaltransformer,guhur2021airbertindomainpretrainingvisionandlanguage,hao2020learninggenericagentvisionandlanguage,qiao2022hophistoryandorderawarepretraining,fried2018speakerfollowermodelsvisionandlanguagenavigation} or object-centric features \cite{an2021neighborviewenhancedmodelvision,chen2022thinkglobalactlocal,moudgil2021soatsceneobjectawaretransformer,qiao2023marchchatinteractiveprompting,zhu2020visionlanguagenavigationselfsupervisedauxiliary}. As Figure~\ref{fig:motivation} demonstrates through a directional selection scenario where only candidate 3 aligns with the instruction, scene-level approaches encoded panoramic views as unitary descriptors via pretrained vision models, while object-based methods decomposed environments into discrete entity features. However, these approaches created two distinct failure modes: (1) holistic features introduced by the global scene representation lack discriminative power in semantically homogeneous environments, e.g., differentiating "bedroom" for candidates 1 and 3, while (2) object-level features overlook contextual relationships critical for navigation, e.g., "window" and "mirror" with similar object-level features, proving insufficient as both candidates share common features. Furthermore, both paradigms suffered from RNN-based history compression \cite{wang2019reinforcedcrossmodalmatchingselfsupervised,sur2019crurcoupledrecurrentunitunification,yan2019hornethierarchicaloffshootrecurrent,chen2014learningrecurrentvisualrepresentation,wu2016imagecaptioningvisualquestion,liu2016recurrentimagecaptionerdescribing}, which may lead to the loss of the temporal information. Specifically, the fixed-size representations used by RNNs may cause additional information loss, particularly in long-horizon navigation tasks \cite{zhang2025crestescalablemaplessnavigation}. This bottleneck hindered the agent's ability to leverage rich temporal context, resulting in suboptimal navigation performance, especially in environments with complex layouts or ambiguous instructions.

% 第三段讲为了克服传统做法中提到的问题（二选一），现在的模型是如何结合这两类信息的，不同的结合方法分别有什么问题（例如问题1，2，3）
Recognizing these limitations, recent advancements \cite{li2023kermknowledgeenhancedreasoning,qian2024recastinggenericpretrainedvision,zhou2021rethinkingspatialrouteprior,yu2025prnet,qiao2022hophistoryandorderawarepretraining,chen2022thinkglobalactlocal,chen2023historyawaremultimodaltransformer,hong2020languagevisualentityrelationship,wang2021structuredscenememoryvisionlanguage,cancelli2025static,gopinathan2025toward,feng2024attention} have attempted hybrid solutions by integrating global and object features. For instance,  transformer-based methods \cite{qiao2022hophistoryandorderawarepretraining,chen2022thinkglobalactlocal,chen2023historyawaremultimodaltransformer} preserve variable-length histories and integrate topological mapping, e.g., DUET \cite{chen2022thinkglobalactlocal}), which improve the navigation performance through external knowledge but at increased computational cost. Advanced architectures \cite{hong2020languagevisualentityrelationship,wang2021structuredscenememoryvisionlanguage} better integrate language, observations, and history for multi-modal reasoning, but performance still degrades with longer instructions due to growing path complexity. However, the following issues still need to be addressed for the current methods. 
First, static feature weighting fails to adapt to varying instruction demands~\cite{cancelli2025static}, e.g., prioritizing objects when hearing "Find the red mug" but scenes for "Enter the sunlit lounge". 
Second, treating all feature levels equally ignores the human-like reasoning hierarchy~\cite{gopinathan2025toward} where low-level cues (object shapes) inform mid-level concepts (furniture arrangements) that scaffold high-level semantics (room purposes).
Third, most hybrid models still rely on RNNs or vanilla transformers for history encoding~\cite{feng2024attention}, causing performance drops in long-horizon SOON~\cite{zhu2021soonscenarioorientedobject} tasks compared to short R2R~\cite{anderson2018visionandlanguagenavigationinterpretingvisuallygrounded} trajectories, which is inadequate for long-horizon tasks. Therefore, these hybrid approaches cannot effectively combine low-level sensory cues with high-level cognitive signals to mimic human-like reasoning.

% 第四段引出，为了克服问题1，我们做了xxx，为了克服问题2，我们做了xxx，为了克服问题3，我们做了xxx

To overcome these challenges, we propose the Multi-level Fusion and Reasoning Architecture (MFRA). 
Firstly, to address the inflexibility of static feature weighting mechanisms in adapting to diverse instructional requirements, MFRA introduces a hierarchical fusion mechanism that efficiently aggregates features across multiple levels of abstraction, from low-level visual cues to high-level semantic concepts. 
Secondly, to deal with the neglect of hierarchical cognitive processing that progressively integrates object-level attributes with spatial configurations and semantic contexts, we design a dynamic reasoning module that leverages instruction-guided attention to align object features with linguistic context, which benefits action prediction in challenging scenarios. 
Third, to mitigate the limitations of conventional temporal encoding architectures in maintaining coherent historical representations for extended navigation sequences, MFRA employs contrastive learning, which enforces similarity between related historical states while distinguishing irrelevant ones by projecting multi-modal input into a shared embedding space, thus enabling seamless integration of visual, linguistic, and temporal modalities. 

% As rich multi-modal information is processed through the fusion and reasoning components, the correlations among visual, linguistic, and temporal signals are effectively learned. The learned representations are able to benefit navigation decision-making, especially for tasks requiring sophisticated scene understanding.
% As humans navigate by hierarchically integrating sensory and cognitive signals \cite{ofner2018hybridactiveinference}, it is essential to develop architectures that fuse multi-level visual-linguistic-temporal features for VLN tasks. 

% 第五段总结我们的模型的优点（就是能做到什么其他模型做不到的好处）以及列出contribution
In this work, we introduce the MFRA, a novel framework designed to enhance embodied agents' reasoning capabilities by effectively integrating visual observations, language instructions, and navigation history. 
The architecture employs a hierarchical fusion mechanism that systematically combines multi-level features spanning from low-level visual cues to high-level semantic concepts across different modalities through \textbf{thinking like a human}. 
Building upon these fused representations, MFRA incorporates a sophisticated reasoning module that utilizes instruction-guided attention and dynamic context integration to infer optimal navigation actions to \textbf{act like a human}.
Through its ability to selectively capture and combine the most relevant visual, linguistic, and temporal signals, MFRA achieves superior decision-making accuracy in complex navigation scenarios, as demonstrated through extensive experimental validation on standard benchmarks. 
% We conduct the experiments on three VLN datasets, i.e.,the REVERIE \cite{qi2020reverieremoteembodiedvisual}, SOON \cite{zhu2021soonscenarioorientedobject}, and R2R \cite{anderson2018visionandlanguagenavigationinterpretingvisuallygrounded}. Our approach outperforms state-of-the-art methods on all splits of these datasets under most metrics. The further experimental analysis demonstrates the effectiveness of our method. 
In summary, we make the following contributions:

\begin{itemize}
    \item We present a novel hierarchical fusion mechanism that systematically combines multi-level features to address the lack of discriminative power in global representations.
    \item We develop a dynamic reasoning module that enhances cross-modal alignment through instruction-guided attention to mitigate the limitations of object-centric approaches.
    \item We conduct extensive experiments to validate the effectiveness of our method and show that it outperforms existing methods with a better generalization ability.
\end{itemize}

\section{Related Work}
{\bfseries Vision-and-Language Navigation.}
Vision-and-Language Navigation (VLN) \cite{anderson2018visionandlanguagenavigationinterpretingvisuallygrounded,chen2022learningunlabeled3denvironments,chen2022thinkglobalactlocal,hong2020languagevisualentityrelationship,qiao2023marchchatinteractiveprompting,tan2019learningnavigateunseenenvironments,wang2021structuredscenememoryvisionlanguage} has emerged as a pivotal research domain due to its transformative potential in applications ranging from healthcare robotics to personalized assistive systems. The core challenge requires agents to interpret diverse natural language instructions including step-by-step directives \cite{wang2019reinforcedcrossmodalmatchingselfsupervised,ku2020roomacrossroommultilingualvisionandlanguagenavigation}, high-level goals \cite{qiao2022hophistoryandorderawarepretraining,zhu2021soonscenarioorientedobject}, and interactive dialogues \cite{qi2020reverieremoteembodiedvisual} and translate them into precise navigation trajectories. Early methodologies \cite{hao2020learninggenericagentvisionandlanguage,wang2019reinforcedcrossmodalmatchingselfsupervised} relied on Recurrent Neural Networks (RNNs) to compress historical observations and actions into fixed-dimensional state vectors, a paradigm later enhanced through richer visual encoding. Notable advancements include scene-level and object-level relational modeling \cite{hong2020languagevisualentityrelationship} and topological maps \cite{chaplot2020neuraltopologicalslamvisual} for long-term spatial memory.

The advent of Vision-and-Language Pretraining (VLP) \cite{radford2021learningtransferablevisualmodels,tan2019lxmertlearningcrossmodalityencoder,zhao2023mindgapimprovingsuccess} catalyzed a paradigm shift toward Transformer-based architectures. Pioneering works like MiniVLN \cite{zhu2024minivlnefficientvisionandlanguagenavigation} introduced a two-stage distillation model covering both the pre-training and fine-tuning phases, while AirBERT \cite{guhur2021airbertindomainpretrainingvisionandlanguage} scaled up multi-modal training data to strengthen modality interactions and RecBERT \cite{li2021recbert} utilizes the [CLS] token within the transformer as a recurrent state to record navigation history. Subsequent innovations refined cross-modal fusion: VLNBERT \cite{hong2021recurrentvisionandlanguagebertnavigation} incorporated recurrent temporal modeling, HAMT \cite{chen2023historyawaremultimodaltransformer} unified language-history-observation sequences via cross-modal attention, and DUET \cite{chen2022thinkglobalactlocal} pioneered graph transformers to bridge local-global spatial reasoning. In this work, we leverage a hierarchical fusion mechanism that aggregates multi-level features across multiple modalities to improve the generalization ability and facilitate the alignment between vision and language for VLN.

\vspace{\baselineskip}
\noindent \textbf{Vision-and-Language Models with Fused Representations.}
The release of the Room-to-Room (R2R) benchmark \cite{anderson2018visionandlanguagenavigationinterpretingvisuallygrounded} marked a pivotal advancement in VLN research, spurring the development of numerous computational models tailored for discrete environments. Initial approaches, exemplified by the Seq2Seq architecture \cite{anderson2018visionandlanguagenavigationinterpretingvisuallygrounded} and the RCM framework \cite{chaplot2020neuraltopologicalslamvisual}, primarily employed imitation learning and reinforcement learning paradigms within standard egocentric visual settings. Subsequent methodological innovations, including CLIP-ViL \cite{shen2021clipbenefitvisionandlanguagetasks}, integrated enhanced visual representations derived from large-scale pretrained models such as CLIP \cite{radford2021learningtransferablevisualmodels,tan2019lxmertlearningcrossmodalityencoder}.  

A significant research trajectory subsequently emerged around the efficient encoding of temporal context, with architectures like VLN-BERT \cite{hong2021recurrentvisionandlanguagebertnavigation} adopting recurrent transformer mechanisms. More recent work \cite{chen2023historyawaremultimodaltransformer,an2023bevbertmultimodalmappretraining} has investigated the incorporation of structured spatial representations, including topological and metric maps, to augment navigational context. Concurrent with these architectural developments, efforts such as ScaleVLN \cite{wang2023scalingdatagenerationvisionandlanguage} have sought to address data scarcity through large-scale training corpus expansion.  

Most recently, the field has witnessed a paradigm shift toward leveraging Large Language Models (LLMs) for VLN, as evidenced by  \cite{zhou2023navgptexplicitreasoningvisionandlanguage,long2023discussmovingvisuallanguage,chen2023a2navactionawarezeroshotrobot,lin2025navcotboostingllmbasedvisionandlanguage}. In this work, we design a reasoning module that leverages hierarchical fused representations to infer navigation actions through instruction-guided attention and dynamic context integration to benefit VLN tasks.
% As noted in the introduction, the ``\verb|acmart|'' document class can
% be used to prepare many different kinds of documentation --- a
% double-anonymous initial submission of a full-length technical paper, a
% two-page SIGGRAPH Emerging Technologies abstract, a ``camera-ready''
% journal article, a SIGCHI Extended Abstract, and more --- all by
% selecting the appropriate {\itshape template style} and {\itshape
%   template parameters}.

% This document will explain the major features of the document
% class. For further information, the {\itshape \LaTeX\ User's Guide} is
% available from
% \url{https://www.acm.org/publications/proceedings-template}.

%   \includegraphics[width=\textwidth]{22.png}
%   \caption{The overall pipeline. (a) The baseline method uses a dual-scale graph transformer to encode the panoramic view, the topological map, and the instruction for action prediction. (b) Our approach incorporates retrieved facts as additional input. The representations of each candidate view are obtained with the purification module, the fact-aware interaction module, and the instruction-guided aggregation module. Best viewed in color.}

\begin{figure*}[t]
    \centering
    \includegraphics[width=\textwidth]{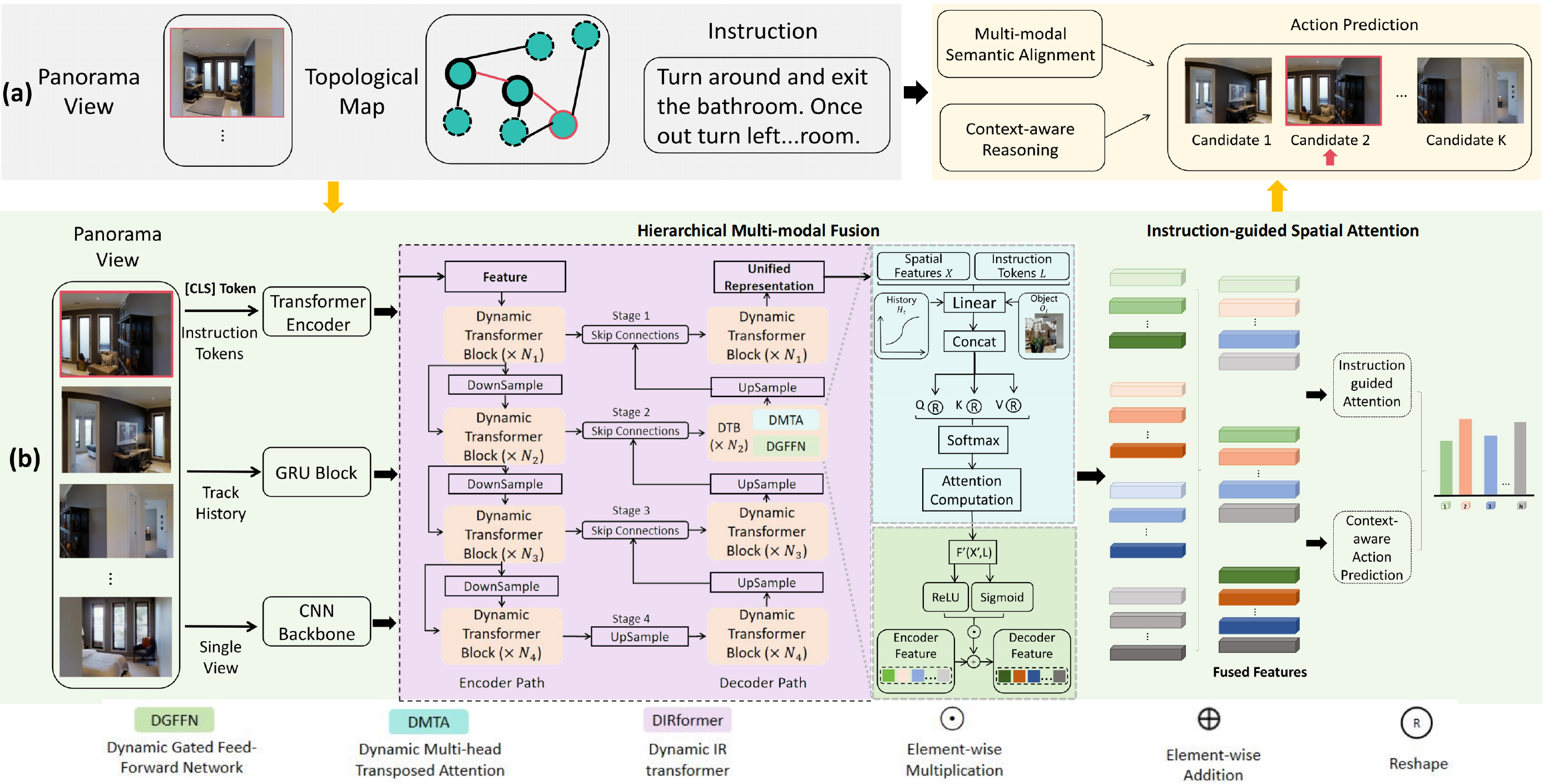}
    \caption{The overall pipeline. (a) The baseline method uses a dual-scale graph transformer to encode the panoramic view, the topological map, and the instruction for action prediction. (b) Our approach incorporates multi-level feature fusion as model input. The representations of each candidate view are obtained with the hierarchical multi-modal fusion, the instruction-guided spatial attention module and the context-aware interaction module. Best viewed in color.}
    \label{fig:wide_image}
\end{figure*}

\section{Methodology}
In this work, we propose a novel and comprehensive framework, termed \textbf{Multi-level Fusion and Reasoning Architecture (MFRA)}, to address the challenges of multi-modal understanding in VLN tasks. VLN requires embodied agents to perceive their visual surroundings, comprehend natural language instructions, and reason over their navigation history to make informed decisions in complex environments. However, effectively integrating these heterogeneous modalities—particularly under ambiguous semantics and dynamic scenes—remains a non-trivial problem. To this end, MFRA is designed to perform hierarchical fusion and semantic reasoning across visual, linguistic, and temporal dimensions. By constructing a unified representation space through multi-level cross-modal interactions, the proposed architecture enables the agent to better align instruction semantics with visual observations and contextual history. This section formally defines the VLN problem setting and elaborates on the key components and mechanisms of MFRA in detail.

\vspace{\baselineskip}
\noindent \textbf{Problem Formulation.}
% We consider the VLN task in a discrete environment, 
The VLN task is formally modeled as a topological graph \( \mathcal{G} = (\mathcal{V}, \mathcal{E}) \), where each node \( v \in \mathcal{V} \) denotes a navigable viewpoint and each edge \( e \in \mathcal{E} \) encodes the connectivity between adjacent viewpoints. The navigation episode begins at a source node \( v_s \), and the agent is instructed to reach a target node \( v_g \) by following a natural language instruction \( I = \{w_1, w_2, ..., w_T\} \), where \( w_t \) represents the \( t \)-th token in the instruction sequence.
At each time step \( t \), the agent receives a panoramic observation \( S_t = \{s_i^t\}_{i=1}^{36} \), consisting of 36 single-view images evenly distributed across the viewing sphere. Each view \( s_i^t \) is represented by a visual feature vector \( v_i^t \in \mathbb{R}^{d_v} \) along with an associated orientation embedding \( o_i^t \in \mathbb{R}^{d_o} \), capturing the spatial direction of the view. For instruction settings involving object grounding, an additional set of object-level features is provided as \( O_t = \{o_j^t\}_{j=1}^{N} \), where each \( o_j^t \) corresponds to a detected or annotated object within the current panorama.
The objective is to learn a navigation policy \( \pi(a_t \mid S_t, I, H_t) \) that predicts an action \( a_t \in \mathcal{A} \) at each timestep, conditioned on the current observation \( S_t \), the instruction \( I \), and the accumulated navigation history \( H_t \). The agent continues to interact with the environment until it selects a termination action or reaches a predefined maximum number of steps.
 
\subsection{Overview of MFRA}
% \vspace{\baselineskip}
% \noindent \textbf{Overview of MFRA.}
To address the challenges of multi-modal semantic alignment and context-aware reasoning in VLN, we propose the \textbf{Multi-level Fusion and Reasoning Architecture (MFRA)}. MFRA is designed to jointly model panoramic visual observations, natural language instructions, and the agent’s navigation history through a hierarchical fusion pipeline and a dynamic reasoning mechanism.

The overall architecture of MFRA consists of three components: (1) \textit{Multi-level Feature Extraction}, which encodes raw visual inputs, instruction tokens, and trajectory history into structured and semantically aligned representations; (2) \textit{Hierarchical Multi-modal Fusion}, which leverages a DIRformer-based structure to align and integrate information across modalities and semantic levels; and (3) \textit{Dynamic Reasoning Module}, which performs instruction-guided attention and context-aware decision making for action prediction.

In the first stage, we utilize a pretrained multi-modal model, CLIP~\cite{radford2021learning}, to obtain aligned visual-language features for both panoramic views and instruction tokens. % Specifically, each single-view image in the panorama is encoded using CLIP’s visual encoder, while the instruction is tokenized and processed by CLIP’s text encoder to obtain a shared embedding space. These features serve as semantically grounded initial representations that facilitate effective cross-modal fusion in subsequent stages. Additionally, object-level features (if available) and historical trajectory embeddings are extracted to support spatial-temporal reasoning.
These multi-level features are then fed into a multi-stage Transformer backbone based on the Dynamic IR Transformer (DIRformer), which enables fine-grained visual-language interaction across different abstraction levels. 
The fused representation is passed to a reasoning module that adaptively integrates temporal context and instruction relevance to infer the most appropriate navigation action.
% By combining CLIP-based aligned representation with hierarchical dynamic fusion and semantic reasoning, MFRA captures both local visual details and global instruction intent, offering enhanced robustness and flexibility for navigating complex, real-world environments.

\subsection{Multi-modal Feature Extraction}
% The agent is required to perceive its environment, interpret natural language instructions, and integrate temporal context to make sequential decisions in VLN tasks. To enable effective downstream fusion and reasoning, we extract semantically aligned representations from three modalities: vision, language, and navigation history.
To enable effective fusion and reasoning, we extract semantically aligned representations from three modalities: vision, language, and navigation history.

At each timestep \( t \), the agent receives a \textbf{panoramic visual observation} composed of 36 discrete single-view images \( S_t = \{s_i^t\}_{i=1}^{36} \). Each view \( s_i^t \) is encoded using the vision encoder of CLIP~\cite{radford2021learning}, producing a set of visual feature vectors \( V_t = \{v_i^t\}_{i=1}^{36} \), where \( v_i^t \in \mathbb{R}^{d_v} \), which are embedded in a shared vision-language space. %, enabling cross-modal alignment in the subsequent fusion stage.

The \textbf{natural language instruction}, represented as a token sequence \( I = \{w_1, w_2, ..., w_T\} \), is embedded using the CLIP text encoder. This yields token-level embeddings \( L = \{l_j\}_{j=1}^{T} \), along with a sentence-level representation \( l_{\text{cls}} \in \mathbb{R}^{d_l} \) that captures global semantic intent. The visual and linguistic embeddings produced by CLIP are inherently aligned in a joint embedding space. %, which facilitates efficient cross-modal interaction.

For scenarios involving object-level grounding, such as in the REVERIE and SOON datasets, we further extract \textbf{object-centric region} features \( O_t = \{o_j^t\}_{j=1}^{N} \) from each panoramic observation. These features are derived from region proposals generated by a pretrained object detector, e.g., Faster R-CNN \cite{ren2016fasterrcnnrealtimeobject}, and are projected into the same embedding space via a learnable adapter to maintain representational consistency with CLIP.

To capture the temporal evolution of the navigation process, we maintain a recurrent history embedding \( H_t \), which summarizes the trajectory observed up to time \( t \). At each step, the current visual observation and its instruction-attended context are fused through a learnable function \( \phi(V_t, L) \), and the result is passed into a Gated Recurrent Unit (GRU) \cite{chung2014empiricalevaluationgatedrecurrent} to update the temporal state:  
\[
h_t = \text{GRU}(h_{t-1}, \phi(V_t, L)).
\]
The resulting representations \( V_t \), \( L \), \( O_t \), and \( H_t \) serve as the foundational multi-modal features, which are then fed into the hierarchical fusion module for joint representation learning and downstream navigation decision-making.

\vspace{\baselineskip}
\noindent \textbf{Language Encoding.}
% \paragraph{Language Encoding.}  
We encode the instruction $I$ using a Transformer encoder to obtain token-level features:
\[
L = \text{Transformer}_{\text{Lang}}(I) = \{l_1, l_2, ..., l_T\}, \quad l_i \in \mathbb{R}^{d}
\]
The first token's embedding $l_{\text{cls}}$ is used as the global instruction context.

\noindent \textbf{Visual Encoding.}
% \paragraph{Visual Encoding.}  
Each single view $s_i^t$ is encoded via a CNN backbone and concatenated with its orientation:
\[
v_i^t = \text{CNN}(s_i^t), \quad x_i^t = [v_i^t; o_i^t]
\]
We denote $X_t = \{x_i^t\}_{i=1}^{36}$ as the full panoramic feature set.

\noindent \textbf{History Encoding.}
% \paragraph{History Encoding.}  
The trajectory history is modeled using a GRU:
\[
h_t = \text{GRU}(x_{1:t}), \quad H_t = \{h_i\}_{i=1}^{t}
\]

\subsection{Hierarchical Multi-modal Fusion}
%\textcolor{red}{The analogy in the introduction section is very good: "treating all feature levels equally ignores the human-like reasoning hierarchy, where low-level cues (object shapes) inform mid-level concepts (furniture arrangements) that scaffold high-level semantics (room purposes)." Please still follow the same logic to formulate this subsection, namely, how to leverage low-level, mid-level, and high-level features, and how they collaborate with each other.combine low-level sensory cues with high-level cognitive signals to mimic human-like reasoning.}
To resolve the inadequacy of uniform multi-modal feature integration, we emulate the human-like cognitive hierarchy, where low-level sensory cues progressively inform higher-order abstractions through designing a hierarchical fusion strategy inspired by DIRformer~\cite{10.1145/3703632} across three semantic tiers: low-level (object shapes), mid-level (spatial arrangements), and high-level (semantic contexts). It enables fine-grained cross-modal alignment at multiple semantic levels and ensures each tier scaffolds the next through guided interactions.

Given the multi-modal features extracted in the previous stage, namely the visual embeddings \( V_t = \{v_i^t\} \), instruction token embeddings \( L = \{l_j\} \), object-level features \( O_t = \{o_j^t\} \), and the temporal history representation \( H_t \), we construct a multi-stage encoder-decoder architecture with dynamic interaction between modalities at each level.
The fusion module is composed of four hierarchical stages. In the encoder path, each stage consists of a series of dynamic Transformer blocks, where each block includes a \textit{Dynamic Multi-head Transposed Attention (DMTA)} layer followed by a \textit{Dynamic Gated Feed-Forward Network (DGFFN)}. At stage \( s \), the feature map \( Z^{(s)} \) is updated as:
\[
Z^{(s)} = \text{DGFFN}\left( \text{DMTA}(Z^{(s-1)}, L) \right),
\]
where the DMTA module performs attention between spatial features and instruction tokens, and DGFFN controls the flow of information through dynamic gating~\cite{sun2025noise,liu2025unifiedvirtualmixtureofexpertsframeworkenhanced}.

\vspace{\baselineskip}
\noindent \textbf{DMTA Module.}
The DMTA module projects the input spatial features \( X \) and instruction tokens \( L \) into query, key, and value spaces via linear transformations to leverage low-level features:
\[
Q = W_q X, \quad K = W_k L, \quad V = W_v L,
\]
and computes attention in a transposed manner:
\[
A = \text{softmax}\left( \frac{Q K^\top}{\sqrt{d}} \right), \quad \tilde{X} = A V.
\]
The attended features \( \tilde{X} \) are added to the input via residual connection, allowing instruction-guided semantic filtering at each scale.

\vspace{\baselineskip}
\noindent \textbf{DGFFN Module.}
The DGFFN module then applies non-linear transformation and dynamic gating to leverage mid-level features to aggregate object interactions (e.g., "sofa adjacent to table") into cohesive arrangements:

\[
F_1 = \text{ReLU}(W_1 X'), \quad F_2 = \sigma(W_2 X'), \quad \text{DGFFN}(X') = F_1 \odot F_2,
\]
where \( \odot \) denotes element-wise multiplication. This allows selective activation of spatial-semantic patterns relevant to the current instruction and visual context.

Following each encoder stage, spatial resolution is reduced by downsampling (e.g. strided convolution), while channel dimensions are expanded to enable broader semantic abstraction. In the decoder path, a symmetric set of upsampling blocks reconstructs higher-resolution representations, with skip connections bridging encoder and decoder stages. The decoder feature at stage \( s \) is computed as:
\[
\hat{Z}^{(s)} = \text{DGFFN}\left( \text{DMTA}(\text{Up}(\hat{Z}^{(s+1)}), L) \oplus Z^{(s)} \right),
\]
where \( \oplus \) denotes channel-wise concatenation. This structure ensures that both fine-grained and global cues are preserved throughout the decoding process. High-level reasoning is also performed by mid-level layouts and low-level features.

To further enhance high-level grounding and temporal reasoning, we incorporate object-level features \( O_t \) and trajectory history \( H_t \) into each fusion stage. These features are projected into token embeddings and fused via auxiliary DMTA layers:
\[
\tilde{X}_o = \text{DMTA}(X, O_t), \quad \tilde{X}_h = \text{DMTA}(X, H_t),
\]
with the combined output:
\[
X'' = X' + \tilde{X}_o + \tilde{X}_h.
\]
This enables the model to reason over fine-grained entities and sequential patterns jointly with visual and linguistic cues. In this way, the tiers achieve a bidirectional interaction: low-level details constrain mid-level arrangements, which in turn contextualize high-level goals. Conversely, high-level semantics (e.g., "find a bedroom") prune irrelevant mid-level configurations, which refine low-level feature extraction. This mimics human top-down and bottom-up reasoning, yielding a unified representation where all tiers co-evolve to resolve ambiguities and align with task objectives.

\begin{algorithm}[ht]
\DontPrintSemicolon
\SetAlgoLined
\SetKwInput{KwInput}{Input}
\SetKwInput{KwOutput}{Output}

\KwInput{Panoramic visual observations $S_t = \{s_i^t\}_{i=1}^{36}$, instruction tokens $I = \{w_1, ..., w_T\}$, optional object features $O_t$, previous trajectory history $H_{t-1}$}
\KwOutput{Predicted navigation action $a_t$}

% Feature Extraction
\textbf{1. Multi-modal Feature Extraction:} \\
\Indp
Encode $S_t$ and $I$ into visual features $V_t = \{v_i^t\}$ using CLIP visual encoder \\
Encode $I$ into token embeddings $L = \{l_j\}$ and global context $l_{\text{cls}}$ using CLIP text encoder \\
If object-level features exist, extract $O_t = \{o_j^t\}$ and project into shared space \\
Fuse $V_t$ and $L$ into context vector $\phi(V_t, L)$ \\
Update temporal history embedding $h_t = \mathrm{GRU}(H_{t-1}, \phi(V_t, L))$ \\
\Indm

% Hierarchical Fusion
\textbf{2. Hierarchical Multi-modal Fusion (DIRformer):} \\
\Indp
Initialize fused representation $Z^{(0)} = V_t$ \\
\For{each encoder stage $s = 1$ to $S$}{
    $Z^{(s)} \gets \mathrm{DGFFN}(\mathrm{DMTA}(Z^{(s-1)}, L))$
}
Integrate object and history context: \\
$\tilde{X}_o = \mathrm{DMTA}(Z^{(S)}, O_t), \quad \tilde{X}_h = \mathrm{DMTA}(Z^{(S)}, H_t)$ \\
$Z_{\text{fused}} = Z^{(S)} + \tilde{X}_o + \tilde{X}_h$ \\
\Indm

% Reasoning and Action
\textbf{3. Dynamic Reasoning and Action Prediction:} \\
\Indp
Compute instruction-guided spatial attention: \\
$\bar{z}_t = \mathrm{Attention}(Z_{\text{fused}}, l_{\text{cls}})$ \\
Form decision vector: $z_t^{\text{final}} = \mathrm{FFN}([\bar{z}_t; h_t])$ \\
Predict action: \\
$a_t = \arg\max_{a_k \in \mathcal{A}_t} \; \mathrm{sim}(z_t^{\text{final}}, v_k)$ \\
\Indm

\KwRet{$a_t$}
\caption{Pseudo-code of MFRA at Time Step $t$}
\label{alg:mfra}
\end{algorithm}

\subsection{Dynamic Reasoning Module}

After hierarchical multi-modal fusion, the agent obtains a unified representation that encodes spatial visual information, instruction semantics, object cues, and temporal trajectory context. To translate this rich representation into actionable decisions, we design a \textbf{Dynamic Reasoning Module} that performs instruction-guided attention and context-aware action prediction.

At each navigation step \( t \), the fused representation from the decoder output of the DIRformer module is denoted as \( Z_t \in \mathbb{R}^{H \times W \times C} \), where each token corresponds to a spatial location in the agent's panoramic view. To allow the agent to focus on regions that are semantically relevant to the instruction, we introduce an \textit{Instruction-guided Spatial Attention} mechanism. Given the global instruction embedding \( l_{\text{cls}} \in \mathbb{R}^{C} \), extracted from the [CLS] token of the CLIP text encoder, we compute attention weights over spatial features as:
\[
\alpha_i = \frac{(W_r z_i)^\top (W_l l_{\text{cls}})}{\sqrt{d}}, \quad \eta = \text{softmax}(\{\alpha_i\}_{i=1}^{HW}), \quad
\bar{z}_t = \sum_{i=1}^{HW} \eta_i z_i,
\]
where \( z_i \) denotes the \( i \)-th token in \( Z_t \), and \( W_r, W_l \) are learnable projection matrices. The resulting vector \( \bar{z}_t \in \mathbb{R}^{C} \) captures the instruction-conditioned visual context at step \( t \).

To incorporate temporal coherence and navigation progress, we further integrate the recurrent history embedding \( h_t \in \mathbb{R}^{C} \), which summarizes all previous observations and actions up to timestep \( t \). The final decision embedding is formed by concatenating the current attended context and historical representation:
\[
z_t^{\text{final}} = \text{FFN}([\bar{z}_t; h_t]),
\]
where FFN denotes a feed-forward network that projects the concatenated vector into a fixed-dimensional decision space.

The action prediction head is a classifier over the candidate navigable directions \( \mathcal{A}_t \), including the stop action. Given the set of candidate view embeddings \( \{v_k\} \subseteq V_t \), we compute their similarity to the decision embedding and apply softmax normalization:
\[
\hat{a}_t = \arg\max_{a_k \in \mathcal{A}_t} \; \text{softmax}\left( \frac{(W_d z_t^{\text{final}})^\top v_k}{\sqrt{d}} \right),
\]
where \( W_d \) is a learnable projection and each \( v_k \) corresponds to the CLIP-encoded feature of a candidate direction. The stop action is modeled similarly, using a dedicated stop embedding vector.

This dynamic reasoning process enables the agent to adaptively attend to instruction-relevant regions, leverage historical context, and evaluate the most appropriate next action in a unified decision space. The module is trained via behavior cloning, minimizing the negative log-likelihood of the expert action at each step.

\subsection{Training Objective}

The proposed MFRA framework is trained in a two-stage manner to fully exploit its multi-modal representation capacity and reasoning capability. The training objective comprises a primary navigation loss and several auxiliary objectives designed to improve semantic grounding, visual-language alignment, and reasoning robustness.

\vspace{\baselineskip}
\noindent \textbf{Supervised Action Learning.}
% \paragraph{Supervised Action Learning.}  
The core objective is to learn a navigation policy that maps the current observation and instruction to the optimal next action. We adopt behavior cloning to supervise the policy network using expert trajectories from the training dataset. Given the ground-truth action sequence \( \{a_t^{*}\}_{t=1}^{T} \), the navigation loss is defined as the cross-entropy between the predicted action distribution and the ground-truth action:
\[
\mathcal{L}_{\text{nav}} = - \sum_{t=1}^{T} \log \pi(a_t^{*} \mid S_t, I, H_t),
\]
where \( \pi(\cdot) \) denotes the learned policy, conditioned on the panoramic observation \( S_t \), the instruction \( I \), and the history embedding \( H_t \).

\vspace{\baselineskip}
\noindent \textbf{Masked Language Modeling (MLM) \cite{devlin2019bertpretrainingdeepbidirectional}.}
% \paragraph{Masked Language Modeling (MLM) \cite{devlin2019bertpretrainingdeepbidirectional}}  
To enhance instruction understanding and facilitate token-level attention, we apply a masked language modeling objective during pretraining. A random subset of instruction tokens is masked and the model is trained to reconstruct them using the fused visual context:
\[
\mathcal{L}_{\text{MLM}} = - \sum_{j \in \mathcal{M}} \log P(w_j \mid S_t, I_{\setminus j}),
\]
where \( \mathcal{M} \) denotes the set of masked positions, and \( I_{\setminus j} \) is the instruction with the \( j \)-th token masked.

\vspace{\baselineskip}
\noindent \textbf{Masked View Classification (MVC) \cite{xie2022simmimsimpleframeworkmasked}.}
% \paragraph{Masked View Classification (MVC) \cite{xie2022simmimsimpleframeworkmasked}}  
To encourage semantic discrimination of visual tokens, we introduce a masked view classification task. A fraction of panoramic views are masked during training, and the model is required to predict their semantic categories:
\[
\mathcal{L}_{\text{MVC}} = - \sum_{i \in \mathcal{V}_m} \log P(c_i \mid v_i),
\]
where \( \mathcal{V}_m \) is the set of masked view indices and \( c_i \) is the target semantic label of view \( v_i \), obtained from an external visual classifier.

\vspace{\baselineskip}
\noindent \textbf{Object Grounding (OG).}
% \paragraph{Object Grounding (OG)}  
For tasks involving object-level grounding, such as REVERIE and SOON, we add an auxiliary loss that encourages the model to locate the correct object referred to by the instruction. The model is trained to predict the correct object \( o^* \) at the final location \( P_D \):
\[
\mathcal{L}_{\text{OG}} = - \log P(o^* \mid I, P_D),
\]

\vspace{\baselineskip}
\noindent \textbf{Overall Loss.}
% \paragraph{Overall Loss.}  
The total training objective is a weighted combination of the above components:
\[
\mathcal{L}_{\text{total}} = \mathcal{L}_{\text{nav}} + \lambda_1 \mathcal{L}_{\text{MLM}} + \lambda_2 \mathcal{L}_{\text{MVC}} + \lambda_3 \mathcal{L}_{\text{OG}}, 
\]
where \( \lambda_1, \lambda_2, \lambda_3 \) are hyperparameters that balance the contributions of each auxiliary task. In our experiments, we empirically set \( \lambda_1 = 1.0 \), \( \lambda_2 = 0.5 \), and \( \lambda_3 = 1.0 \).

This joint training objective enables MFRA to learn semantically rich and well-aligned representations across modalities, while simultaneously grounding object references and enhancing instruction-conditioned reasoning.

\begin{figure}[t]
    \centering
    \includegraphics[width=0.47\textwidth]{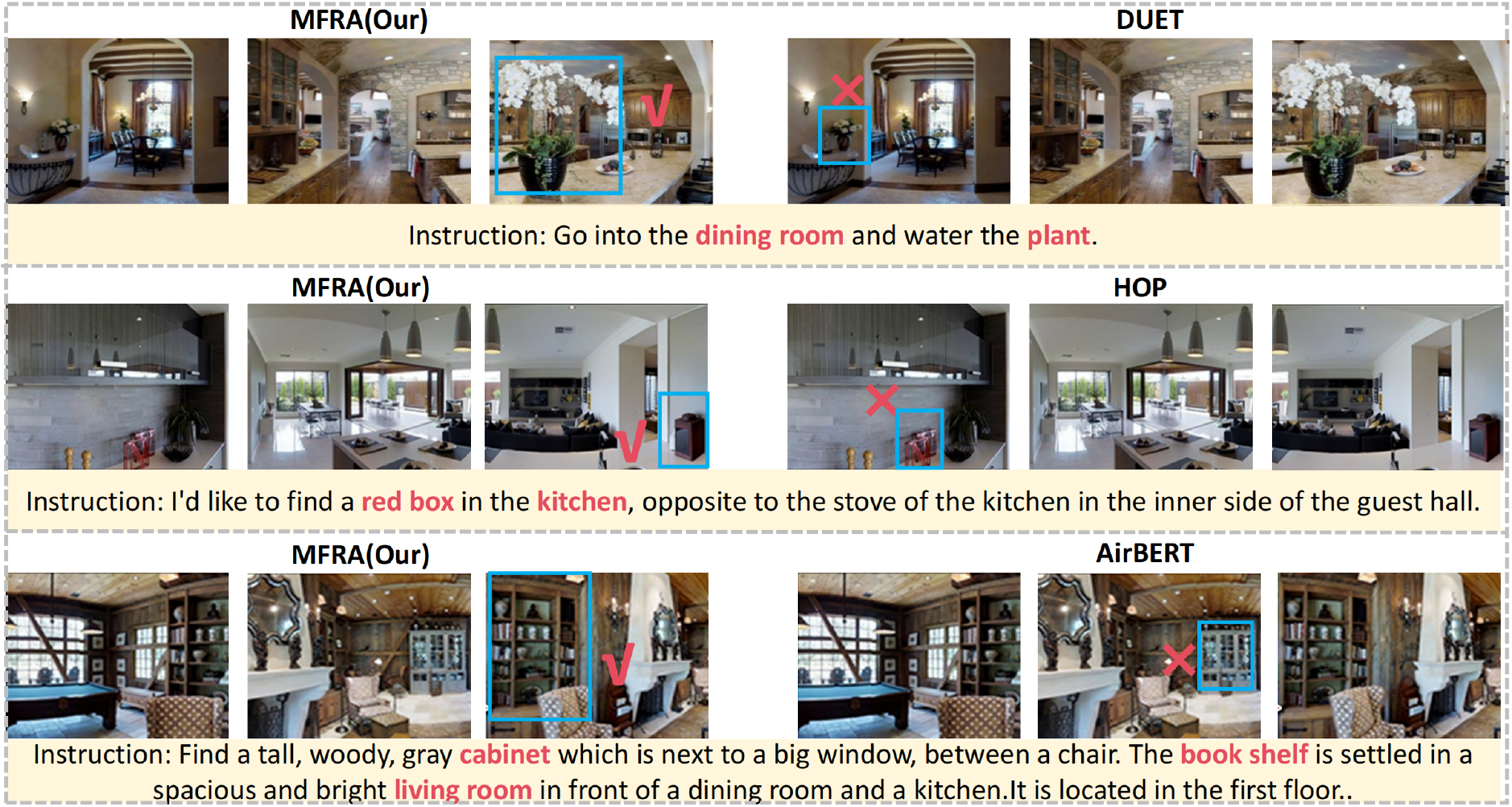}
    \vspace{-0.2cm}
    \caption{Visualization of navigation examples. The sentence within the yellow box is the navigation instruction for the agent. We show a comparison where our MFRA chooses the right location while the baseline model makes the wrong choice. Best viewed in color.}
    \label{fig:wide_image}
\end{figure}

\begin{table*}[t]
\centering
\caption{Performance comparison with SOTA methods on the R2R dataset. All metrics are reported in \%. TL is in meters.}
\vspace{-0.2cm}
\label{tab:reverie_results}
\resizebox{\textwidth}{!}{
\begin{tabular}{l|cccccc|cccccc|cccccc}
\toprule
\multirow{2}{*}{Method} & \multicolumn{6}{c|}{Val Seen} & \multicolumn{6}{c|}{Val Unseen} & \multicolumn{6}{c}{Test Unseen} \\
& TL↓ & OSR↑ & SR↑ & SPL↑ & RGS↑ & RGSPL↑ & TL↓ & OSR↑ & SR↑ & SPL↑ & RGS↑ & RGSPL↑ & TL↓ & OSR↑ & SR↑ & SPL↑ & RGS↑ & RGSPL↑ \\
\midrule
Seq2Seq & 12.88 & 35.70 & 29.59 & 24.01 & 18.97 & 14.96 & 11.07 & 8.07 & 4.20 & 2.84 & 2.16 & 1.63 & 10.89 & 6.88 & 3.99 & 3.09 & 2.00 & 1.58 \\
VLNBERT & 13.44 & 53.90 & 51.79 & 47.96 & 38.23 & 35.61 & 16.78 & 35.02 & 30.67 & 24.90 & 18.77 & 15.27 & 15.68 & 32.91 & 29.61 & 23.99 & 16.50 & 13.51 \\
AirBERT & 15.16 & 49.98 & 47.01 & 42.34 & 32.75 & 30.01 & 18.71 & 34.51 & 27.89 & 21.88 & 18.23 & 14.18 & 17.91 & 34.20 & 30.28 & 23.61 & 16.83 & 13.28 \\
HOP & 13.80 & 54.88 & 53.76 & 47.19 & 38.65 & 33.85 & 16.46 & 36.24 & 31.78 & 26.11 & 18.85 & 15.73 & 16.38 & 33.06 & 30.17 & 24.34 & 17.69 & 14.34 \\
DUET & 13.86 & 73.68 & 71.75 & 63.94 & 57.41 & 51.14 & 22.11 & 51.07 & 46.98 & 33.73 & 32.15 & 23.03 & 21.30 & 56.91 & 52.51 & 36.06 & 31.88 & 22.06 \\
NaviLLM & 14.10 & 75.00 & 73.12 & 66.08 & 59.20 & 53.17 & 22.75 & 53.91 & 48.53 & 34.76 & 33.42 & 23.88 & 20.90 & 57.10 & 51.80 & 37.15 & 30.91 & 21.84 \\
VLN-PETL & 13.92 & 70.26 & 68.87 & 61.55 & 54.30 & 49.78 & 21.64 & 49.70 & 45.32 & 32.44 & 30.99 & 22.07 & 20.58 & 53.68 & 49.63 & 35.27 & 30.12 & 21.10 \\
LaNA & 14.08 & 71.92 & 70.04 & 62.89 & 56.01 & 50.10 & 21.88 & 50.13 & 45.76 & 33.55 & 31.67 & 22.33 & 21.02 & 55.11 & 50.47 & 36.12 & 31.08 & 21.49 \\
ETPNav & 13.71 & 72.84 & 71.01 & 63.00 & 56.39 & 51.25 & 21.93 & 51.32 & 46.21 & 33.80 & 31.21 & 22.17 & 21.10 & 55.43 & 51.06 & 36.04 & 31.26 & 21.73 \\
\textbf{MFRA (Ours)} & \textbf{12.84} & \textbf{79.20} & \textbf{76.88} & \textbf{70.45} & \textbf{61.00} & \textbf{56.07} & \textbf{21.85} & \textbf{55.21} & \textbf{50.44} & \textbf{35.38} & \textbf{34.51} & \textbf{24.45} & \textbf{17.32} & \textbf{57.58} & \textbf{52.43} & \textbf{39.21} & \textbf{32.39} & \textbf{23.64} \\
\bottomrule
\end{tabular}
}
\end{table*}

\section{Experiments}

\subsection{Datasets and Evaluation Metrics}

We evaluate the proposed MFRA model on three widely-used VLN benchmarks: REVERIE~\cite{qi2020reverieremoteembodiedvisual}, SOON~\cite{zhu2021soonscenarioorientedobject}, and R2R~\cite{anderson2018visionandlanguagenavigationinterpretingvisuallygrounded}. These datasets cover a wide range of navigation and grounding challenges with varying instruction complexity and visual ambiguity.

\textbf{REVERIE} provides high-level navigation instructions with an average length of 21 words. Each panoramic node includes predefined object bounding boxes, and the agent is required to identify the target object at the end of the navigation path. Trajectory lengths range from 4 to 7 steps.
\textbf{SOON} features longer and more complex instructions (average length of 47 words) and does not include predefined bounding boxes. The agent must predict the object center location using object detectors~\cite{jain2022bottom}. Path lengths range from 2 to 21 steps.
\textbf{R2R} consists of step-by-step navigation instructions with an average length of 32 words and an average trajectory length of 6 steps. Unlike REVERIE and SOON, R2R focuses solely on spatial navigation without explicit object grounding.

We adopt standard VLN evaluation metrics, including:  
(1) \textbf{Trajectory Length (TL)}: average length of navigation paths;  
(2) \textbf{Navigation Error (NE)}: average distance (in meters) between the agent’s final location and the goal;  
(3) \textbf{Success Rate (SR)}: percentage of episodes where NE is less than 3 meters;  
(4) \textbf{Oracle Success Rate (OSR)}: SR assuming an oracle stop policy;  
(5) \textbf{SPL}: SR weighted by path efficiency.  
For datasets with grounding tasks (REVERIE, SOON), we further report:  
(6) \textbf{Remote Grounding Success (RGS)}: proportion of successfully grounded instructions;  
(7) \textbf{RGSPL}: RGS weighted by path length. For all metrics except TL and NE, higher values indicate better performance.

% \vspace{\baselineskip}
% \noindent \textbf{Model Architecture.}
% \paragraph{Model Architecture.}  
We adopt CLIP-ViT-B/16  to extract aligned visual-language features for both panoramic views and instruction tokens. Object features on REVERIE are obtained using ViT-B/16 pretrained on ImageNet, while object bounding boxes on SOON are extracted using the BUTD detector~\cite{jain2022bottom}. 
The fusion module is implemented using a 4-stage DIRformer encoder-decoder architecture with DMTA and DGFFN blocks. Cross-modal interaction layers are initialized from LXMERT~\cite{tan2019lxmert}.

\subsection{Experimental Results and Analysis}

We compare our proposed MFRA model with a range of representative state-of-the-art methods on the R2R dataset, covering both traditional sequence-based baselines \cite{anderson2018visionandlanguagenavigationinterpretingvisuallygrounded} and recent transformer-based or knowledge-enhanced architectures \cite{hong2021recurrentvisionandlanguagebertnavigation,guhur2021airbertindomainpretrainingvisionandlanguage,qiao2022hophistoryandorderawarepretraining,chen2022thinkglobalactlocal}, as well as newly proposed models such as NaviLLM~\cite{zheng2024learninggeneralistmodelembodied}, VLN-PETL~\cite{qiao2023vlnpetlparameterefficienttransferlearning}, LaNA~\cite{wang2023lanalanguagecapablenavigatorinstruction}, and ETPNav~\cite{an2024etpnavevolvingtopologicalplanning}. The quantitative results on the validation seen, validation unseen, and test unseen splits are summarized in Table~\ref{tab:reverie_results}. The comparison results on other datasets can be found in the supplementary material.

The results presented in Table~\ref{tab:reverie_results} comprehensively demonstrate the superiority of our proposed MFRA framework across all evaluation metrics on the REVERIE dataset. Compared with prior state-of-the-art methods, MFRA achieves consistent and significant improvements in both navigation accuracy and object grounding precision. Specifically, MFRA outperforms DUET by +5.13\% in Success Rate (SR) and +6.51\% in SPL on the validation seen split, while also achieving higher Remote Grounding Success (RGS) and RGSPL, confirming the effectiveness of our hierarchical fusion and semantic reasoning design. Moreover, compared with recent advances such as NaviLLM and ETPNav—which incorporate large language model priors and evolving topological planning—MFRA still shows superior performance, with a clear advantage of 2–5\% on both SR and RGS across most splits. While VLN-PETL demonstrates efficiency through parameter-efficient tuning and LaNA introduces bi-directional language understanding, their navigation accuracy remains lower than that of MFRA, particularly under domain shifts and object ambiguity.

Notably, the performance drop from seen to unseen environments is much smaller than that of all baseline methods, indicating stronger generalization capabilities. This robustness stems from the use of CLIP-based multi-modal features trained with large-scale language supervision, and from our unified DIRformer fusion backbone which supports modality-invariant representation learning. Furthermore, the integration of dynamic multi-head transposed attention and gated feed-forward networks enables our model to selectively attend to instruction-relevant regions, improving cross-modal semantic alignment and decision confidence. Compared to object-centric approaches such as AirBERT or VLNBERT, MFRA benefits from a fact-level grounding mechanism that captures richer contextual information beyond static object labels. This allows the agent to reason over both fine-grained visual cues and high-level semantic references within instructions. Overall, the strong empirical results validate that MFRA is capable of learning robust, generalizable, and semantically grounded navigation policies, outperforming existing methods under varying conditions of instruction complexity, scene diversity, and task ambiguity.

 \subsection{Ablation Study}

To better understand the contribution of each component in our proposed MFRA framework, we conduct a comprehensive ablation study on the REVERIE validation unseen split. Specifically, we progressively remove or modify critical modules in the architecture and evaluate their impact on both navigation and grounding performance. The results are summarized in Table~\ref{tab:ablation}.

As shown in Table~\ref{tab:ablation}, removing the hierarchical DIRformer fusion module leads to the most significant performance degradation, with a 4.42\% drop in SR and over 2.3\% drop in RGSPL, confirming the central role of multi-level cross-modal reasoning in MFRA. Disabling the instruction-guided attention module also reduces SR and RGS, indicating that semantic alignment between instruction-level abstraction and visual context is critical for accurate decision-making. The exclusion of the history-fact interaction module leads to a moderate decline, suggesting that temporal information contributes meaningfully to global context reasoning. Finally, replacing CLIP-based fact features with conventional CNN+LSTM-based representations causes the largest overall performance drop in both navigation and grounding accuracy, demonstrating that pretrained multi-modal embeddings offer superior semantic alignment capabilities. These ablation results collectively verify that each module in MFRA—especially DIRformer fusion and CLIP-based representation—plays a vital role in enabling accurate and generalizable instruction-grounded navigation.

\begin{table}[ht]
\centering
\caption{Ablation study results on the R2R val unseen split.}
\vspace{-0.2cm}
\label{tab:ablation}
\resizebox{0.48\textwidth}{!}{
\begin{tabular}{l|ccccc}
\toprule
Model Variant & SR↑ & SPL↑ & OSR↑ & RGS↑ & RGSPL↑ \\
\midrule
Full MFRA (Ours) & \textbf{50.44} & \textbf{35.38} & \textbf{55.21} & \textbf{34.51} & \textbf{24.45} \\
w/o DIRformer Fusion & 46.02 & 32.01 & 51.87 & 31.08 & 22.13 \\
w/o Instruction-guided Attention & 47.12 & 33.10 & 52.44 & 32.28 & 22.86 \\
w/o History-Fact Interaction & 48.30 & 34.12 & 53.53 & 33.63 & 23.58 \\
w/o CLIP-based Fact Representation & 44.91 & 31.04 & 50.11 & 30.25 & 21.14 \\
\bottomrule
\end{tabular}
}
\end{table}
\vspace{-0.3cm}

\begin{table}[ht]
\centering
\caption{Analysis of multi-modal feature contributions on the R2R val unseen split.}
\vspace{-0.2cm}
\label{tab:modality-analysis}
\resizebox{0.48\textwidth}{!}{
\begin{tabular}{l|ccccc}
\toprule
Feature Configuration & SR↑ & SPL↑ & OSR↑ & RGS↑ & RGSPL↑ \\
\midrule
All Modalities (Full MFRA) & \textbf{50.44} & \textbf{35.38} & \textbf{55.21} & \textbf{34.51} & \textbf{24.45} \\
w/o Visual Feature & 41.27 & 27.65 & 47.04 & 28.12 & 19.90 \\
w/o Language Feature & 36.08 & 23.54 & 43.19 & 25.89 & 17.43 \\
w/o History Feature & 46.11 & 32.79 & 52.44 & 32.15 & 22.04 \\
Visual + Language only & 48.05 & 34.01 & 54.12 & 33.22 & 23.27 \\
Visual + History only & 43.37 & 30.12 & 50.86 & 30.84 & 21.06 \\
Language + History only & 40.90 & 28.23 & 48.21 & 29.14 & 20.36 \\
\bottomrule
\end{tabular}
}
\end{table}

\subsection{Multi-modal Feature Contribution Analysis}

To further investigate the contribution of individual modalities in MFRA, we conduct a series of controlled experiments by selectively removing vision, language, or history components from the fused representation. Specifically, we analyze how navigation performance is affected when one or more modalities are excluded from the multi-modal reasoning process. The results on the REVERIE validation unseen split are reported in Table~\ref{tab:modality-analysis}.

The results demonstrate that all three modalities—vision, language, and history—play complementary roles in supporting effective navigation and grounding. When the visual modality is removed, performance drops sharply across all metrics (-9.17\% SR and -4.55\% RGSPL), indicating that grounded visual perception is essential for scene understanding. Excluding the instruction input leads to the largest performance degradation (-14.36\% SR and -7.02\% ), as language provides high-level semantic guidance that conditions the agent’s decisions. The history feature also contributes meaningfully, particularly in grounding-oriented metrics, as removing it causes a -2.41\% drop in RGSPL. We further analyze partial modality combinations. While visual and language features alone can support reasonable performance (48.05\% SR), integrating history yields additional gains, highlighting the importance of temporal context. In contrast, using only language and history without visual grounding leads to significant performance degradation.

% Overall, this analysis validates our design choice to explicitly model and integrate all three modalities, and confirms that cross-modal reasoning over vision, language, and navigation history is crucial for robust and generalizable VLN performance.

\subsection{Cross-Dataset Evaluation on REVERIE and SOON}

\begin{figure}[ht]
    \centering
    \includegraphics[width=0.48\textwidth]{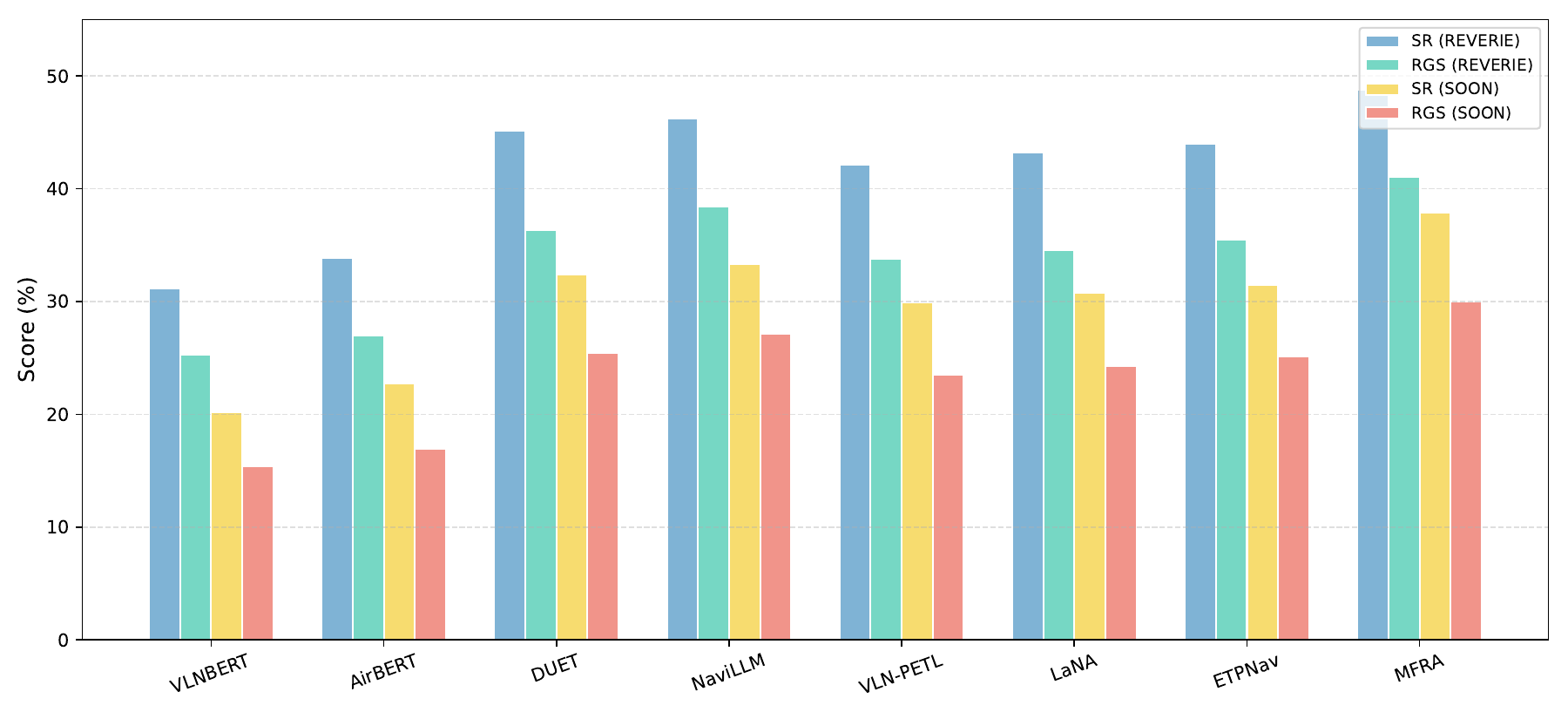}
    \vspace{-0.2cm}
    \caption{Performance comparison on REVERIE and SOON validation unseen splits. MFRA achieves consistent improvements in Success Rate (SR) and Remote Grounding Success (RGS) over baseline methods.}
    \label{fig:cross_dataset_bar}
\end{figure}

To further evaluate the generalization performance of MFRA, we conduct experiments on the REVERIE and SOON datasets, both of which pose greater challenges in object grounding and long-horizon instruction following. These datasets serve as rigorous benchmarks for testing multi-modal fusion, semantic reasoning, and policy robustness in unseen environments.
Figure~\ref{fig:cross_dataset_bar} illustrates the Success Rate (SR) and Remote Grounding Success (RGS) of MFRA in comparison with a broad set of representative baselines, including both classical transformer-based models such as VLNBERT, AirBERT, and DUET, as well as more recent approaches like NaviLLM, VLN-PETL, LaNA, and ETPNav. Across both datasets, MFRA consistently achieves the highest performance, demonstrating stronger instruction-following accuracy and grounding precision under complex and diverse scene conditions.

Notably, while methods such as NaviLLM and ETPNav benefit from large language model priors or topological planning, their performance still lags behind MFRA, particularly in generalizing to previously unseen environments. VLN-PETL and LaNA introduce parameter-efficient adaptation and generative capabilities, but face limitations when dealing with fine-grained object semantics and long instruction spans. In contrast, MFRA leverages unified visual-language representations derived from CLIP, combined with a hierarchical DIRformer-based fusion mechanism, enabling effective cross-modal alignment and robust reasoning over both low-level visual cues and high-level semantic structures.

These results collectively demonstrate the superior generalization ability of MFRA across multiple instruction formats and grounding requirements, validating the effectiveness of its design under realistic embodied navigation scenarios.

\section{Conclusion}
In this paper, we propose MFRA, a multi-level fusion and reasoning architecture that enhances the agent’s ability to reason over visual observations, language instructions and navigation history for VLN. Our work utilizes a hierarchical fusion mechanism that aggregates multi-level features across multiple modalities, ranging from low-level visual cues to high-level semantic concepts. We further design a reasoning module that leverages fused representations to infer navigation actions through instruction-guided attention and dynamic context
integration. We illustrate the good interpretability of MFRA and provide case study in deep insights. Our approach achieves excellent improvement on many VLN tasks, demonstrating that hierarchical fusion and reasoning is a promising direction in improving VLN and Embodied AI. For future work, we will improve our MFRA with larger training data and employ it on VLN in continuous environments.

\twocolumn[  % 开始双栏模式
  % 双栏内容
]
\bibliographystyle{ACM-Reference-Format}

\bibliography{main}

\end{document}